\documentclass[runningheads]{llncs}
\usepackage[T1]{fontenc}
\usepackage{graphicx,verbatim,amsmath,amssymb,booktabs,multirow}
\begin{document}
\title{DeGenseGS: Geometrically and Semantically Decoupled Surgical Scene Understanding in 4D Gaussian Splatting}
%
\author{Yimo Wang\inst{1,3}\thanks{Yimo Wang and Bin Kang contributed equally to this work.} \and
Bin Kang\inst{2}$^{\star}$ \and
Shuojue Yang\inst{3} \and
Yueming Jin\inst{3}\thanks{Corresponding author.}}
%
\authorrunning{Y. Wang et al.}
\institute{School of Automation, Southeast University, Nanjing 210096, China \\
\email{yimowang@seu.edu.cn} \and
School of Internet of Things, Nanjing University of Posts and Telecommunications, Nanjing, China \\
\email{kangb@njupt.edu.cn} \and
Department of Biomedical Engineering, National University of Singapore, Singapore \\
\email{\{yimowang,s.yang\}@u.nus.edu, ymjin@nus.edu.sg}}
  
\maketitle              
\begin{abstract}
Real-time, text-promptable 4D reconstruction is indispensable for autonomous surgical interaction. Severe misalignment between semantic meaning and physical anatomy still persists, largely because existing solutions integrate Vision-Language Models into deformable fields via a rigid coupling scheme that tightly binds semantic features to geometric warping. In this paper, we propose DeGenseGS, Geometrically and Semantically Decoupled Surgical Scene Understanding in 4D Gaussian Splatting, a novel framework that independently models semantic evolution and geometric deformation. Specifically, we propose a HexPlane-based spatiotemporal entanglement module that uses shared kinematic latents to synchronize semantic mutations with scene dynamics, while explicitly disentangling semantic updates from geometric deformation. To further ensure robustness against reconstruction artifacts, we devise a Rasterization-Native Semantic Extraction mechanism that infers semantics from topologically continuous feature maps. Additionally, we incorporate an angular-aligned optimization strategy that conforms to the native hyperspherical latent space, thereby preventing semantic distortion. Extensive evaluations on the CholecSeg8k and EndoVis18 datasets demonstrate that DeGenseGS achieves state-of-the-art performance. Our framework yields enhanced geometric completeness and robust semantic-anatomic alignment, enabling spatially continuous segmentation despite drastic tissue deformation and topological transitions.


\end{abstract}
\section{Introduction}

In the progression towards autonomous Robotic-Assisted Minimally Invasive Surgery (RAMIS), surgical perception systems must evolve from passive geometric observers to active cognitive agents. While high-fidelity 4D reconstruction provides essential spatial mapping, it remains semantically agnostic\cite{ref1}. To enable advanced downstream capabilities, such as safety-critical zone alerting and fine-grained instrument tracking, the spatial reconstruction must be augmented with text-promptable semantic understanding. This requires a framework capable of simultaneously tracking dynamic anatomy in real-time and assigning queryable semantic identities to the reconstructed environment, thereby bridging the gap between low-level geometric sensing and high-level clinical reasoning.

Recent advancements have transitioned surgical scene reconstruction from implicit neural representations \cite{ref2,ref3} to explicit 4D Gaussian Splatting (4DGS) \cite{ref4,ref5,ref6,ref7}, achieving remarkable rendering efficiency and visual fidelity. Concurrently, surgical Vision-Language Models (VLMs)\cite{ref8,ref9,ref10,ref11} have enabled sophisticated text-promptable 2D scene interpretation \cite{ref12}. Despite these parallel successes, a critical dimensionality gap persists: dynamic 4D reconstructions inherently lack semantic awareness, whereas 2D VLMs lack essential spatial grounding \cite{ref1}. While recent pioneering efforts \cite{ref13,ref14,ref15} attempt to bridge this gap by lifting 2D semantic priors into dynamic volumetric fields, establishing robust semantic-geometric alignment remains an open challenge. Specifically, the severe occlusions, complex tissue deformations, and rapid topological changes characteristic of endoscopic environments \cite{ref16,ref17} expose the fragility of these early rigidly coupled representations. This inherent limitation necessitates a specialized framework to explicitly decouple spatial deformation from semantic evolution.

\begin{figure}[t]
    \centering
    \includegraphics[width=\textwidth]{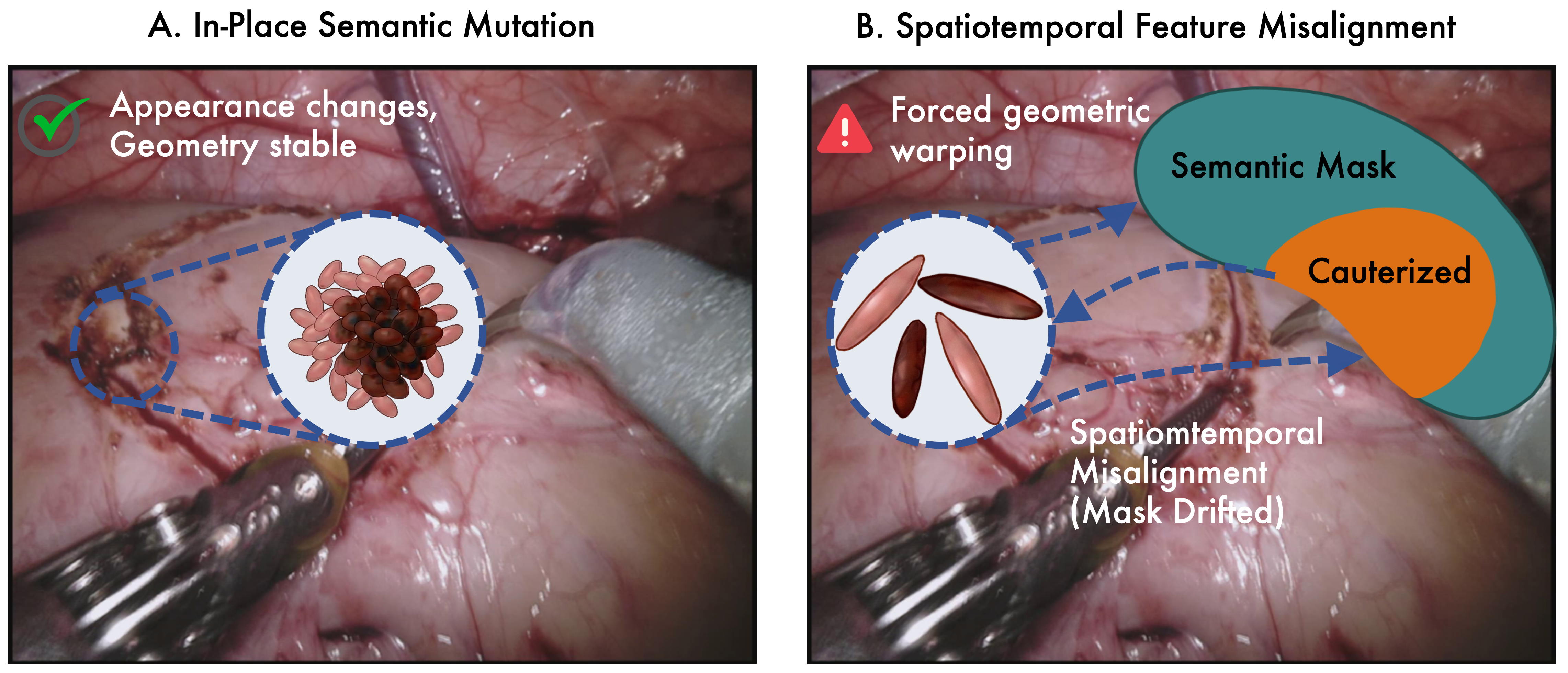}
    \caption{Illustration of spatiotemporal feature misalignment in existing coupled 4D Gaussian Splatting frameworks. \textbf{(A) Physical Reality:} During topology-altering events such as cauterization, tissue appearance mutates significantly while the underlying geometry remains largely stable. \textbf{(B) Limitation of existing Coupled Frameworks:} Due to the gradient sensitivity variation, the coupled framework  leads to unfounded geometric warping (blue arrows) to accommodate the new semantic identity.}
    \label{fig:feature_misalignment}
\end{figure}

To populate an independent semantic field, extracting robust 2D priors is a prerequisite. While SAM-based models yield sharp boundaries, their features lack explicit semantic representation \cite{ref18,ref19}. Accordingly, incorporating Vision-Language Models (VLMs) is indispensable for text-promptable surgical scene understanding. The key challenge lies in simultaneously modeling semantic evolution alongside geometric deformation. Current 3DGS baselines cast these two factors into a single deformation framework, leveraging the Flexible Deformation Model (FDM) that couples spatial warping with semantic refinement in a shared pipeline \cite{ref6,ref7,ref20}. This coupled design suffers from the gradient sensitivity issue \cite{ref21} in surgical events such as surgical cauterization, where tissue appearance mutates significantly but the physical geometry remains largely static. Specifically, we observe that the network gradients of FDM are far more sensitive to semantic shifts than to subtle geometric changes \cite{ref22}. As semantic and geometric representations share the same FDM weight, gradients dominated by semantic changes will become excessively large in traditional solutions, inadvertently leading to spurious geometric warping during the surgical cauterization task. As shown in Fig. \ref{fig:feature_misalignment}, there exist misaligned representations for semantic evolution and geometric deformation
in FDM. These erroneous deformations propagate to the semantic mask, eventually undermining the geometric and semantic consistency of the reconstructed scene.

In this paper, we propose DeGenseGS, namely Geometrically and Semantically Decoupled Surgical Scene Understanding in 4D Gaussian Splatting, which integrates VLM in 4D
Gaussian Splatting for separately model semantic evolution and geometric deformation.
First, at the spatiotemporal level, a HexPlane-based module is proposed to extract shared kinematic latents. These latents are routed to two decoupled decoding branches to independently model geometric deformation and semantic evolution. This design mitigates gradient interference, preventing semantic losses from dominating the optimization of geometric parameters. As a result, tissue appearance can evolve freely without inducing spurious spatial displacements.
Second, to reduce reconstruction artifacts, we propose a Rasterization-Native Semantic Extraction mechanism that decouples semantic features from deformed 4D Gaussian kernels. It performs graph-based analysis on topologically continuous 2D feature maps and leverages RGB renderings as structural guidance to correct local VLM inconsistencies, producing robust, hole-free segmentation masks.
Finally, we introduce a manifold-aligned optimization paradigm for robust semantic grounding, where the semantic distillation is formulated as a metric learning problem.  The contributions of this paper are as follows:
\begin{itemize}
    \item We propose DeGenseGS, the first text-promptable 4DGS framework that explicitly decouples spatial deformation from semantic evolution, enabling fine-grained surgical interaction.
    \item We introduce a Rasterization-Native Semantic Extraction mechanism that leverages RGB structural guidance to rectify VLM inconsistencies and ensure hole-free, precise segmentation boundaries.
    \item We propose a Manifold-Aligned Optimization strategy to regularize the VLM latent space via angular distillation, effectively mitigating feature collapse in dynamic surgical environments.
    \item State-of-the-art performance is validated on comprehensive datasets, demonstrating superior semantic-geometric alignment capability.
\end{itemize}

\section{Methodology}
\begin{figure}[t]

    \centering

    \includegraphics[width=\textwidth]{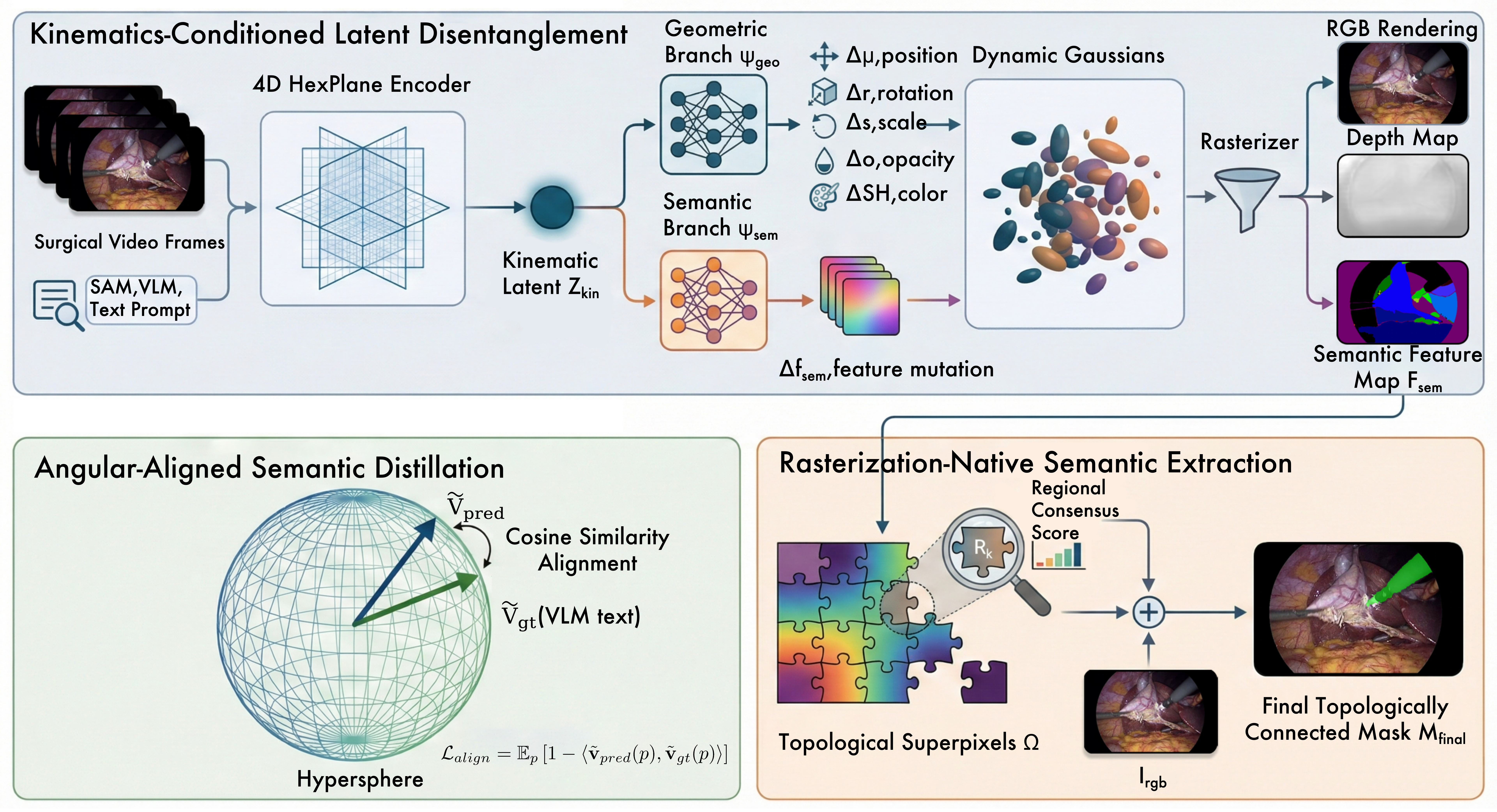}

    \caption{Overview of the proposed DeGenseGS framework. To prevent spurious geometric warping, 
    a HexPlane encoder extracts shared kinematic latents to enable independent geometric and semantic decoding
    , thereby explicitly decoupling the optimization pathways. Subsequently, Angular-Aligned Semantic Distillation, together with a Rasterization-Native Semantic Extraction mechanism, strengthens feature grounding and facilitates accurate boundary recovery from rasterized feature maps.}

    \label{fig:pipeline}

\end{figure}

DeGenseGS is a 4D Gaussian Splatting framework designed for robust surgical scene understanding (Fig. \ref{fig:pipeline}). It explicitly decouples the representation of semantic evolution from spatial deformation using a kinematics-conditioned latent disentanglement mechanism. And it also recovers precise boundaries via a rasterization-native semantic extraction.

\subsection{Preliminaries}
3D Gaussian Splatting (3DGS) represents scenes using anisotropic 3D Gaussians, defined by center $\boldsymbol{\mu} \in \mathbb{R}^3$, covariance $\boldsymbol{\Sigma}$, opacity $\sigma$, and color $c$. To render a novel view, 3D Gaussians are projected into 2D splats and alpha-blended in depth order: $\hat{\mathbf{C}}(\mathbf{p})=\sum_{i=1}^N c_i \alpha_i \prod_{j=1}^{i-1}(1-\alpha_j)$.

\subsection{Kinematics-Conditioned Latent Disentanglement}
Existing frameworks tightly couple semantic features with geometric deformations, causing unnatural spatial warping during in-place semantic mutations (e.g., cauterization). We address this by disentangling semantic evolution from spatial displacement. Given a Gaussian at position $p \in \mathbb{R}^3$ and time $t$, a multi-resolution HexPlane encoder extracts a spatiotemporal feature $\mathbf{h}_{st} = \mathcal{H}(p, t)$. A shared base network processes $\mathbf{h}_{st}$ to yield a local dynamics representation $\mathbf{h}_{base}$. We project $\mathbf{h}_{base}$ into property-specific states to form a \textit{latent kinematic descriptor} $\mathcal{Z}_{kin}$:
\begin{equation}
    \mathcal{Z}_{kin} = \bigoplus_{k \in \mathcal{K}} \phi_k(\mathbf{h}_{base}) \in \mathbb{R}^{d_{kin}}, \quad \mathcal{K} = \{pos, scale, rot, opa\}
\end{equation}
where $\phi_k(\cdot)$ denotes linear transformations. Rather than parameterizing the semantic feature $f_{sem}(t)$ via deformed geometry, two independent branches decode geometric deformations $\Delta \mathcal{G}_{t}$ and semantic updates $\Delta f_{sem, t}$ directly from $\mathcal{Z}_{kin}$:
\begin{equation}
    \Delta \mathcal{G}_{t} = \Psi_{geo}(\mathcal{Z}_{kin}), \quad \Delta f_{sem, t} = \Psi_{sem}(\mathcal{Z}_{kin})
\end{equation}
This conditionally isolates geometric and semantic updates. During topology-altering events with static geometry but changing appearance, $\Psi_{sem}$ freely triggers semantic shifts ($\Delta f_{sem} \neq \mathbf{0}$) without forcing false geometric displacements ($\Delta \mathcal{G} \to \mathbf{0}$).

\subsection{Angular-Aligned Semantic Distillation}
To distill 2D VLM knowledge, we optimize the angular similarity between rendered and ground-truth VLM features within their native hyperspherical space. Features $\mathbf{v}_{pred}(p)$ and $\mathbf{v}_{gt}(p)$ are $L_2$-normalized to $\tilde{\mathbf{v}}$. The alignment loss $\mathcal{L}_{align}$ minimizes their angular discrepancy:
\begin{equation}
    \mathcal{L}_{align} = \mathbb{E}_{p} \left[ 1 - \langle \tilde{\mathbf{v}}_{pred}(p), \tilde{\mathbf{v}}_{gt}(p) \rangle \right]
\end{equation}
To enforce semantic consistency within anatomical structures, we add a region smoothness regularization $\mathcal{L}_{smooth} = \mathbb{E}_{R \in \Omega} [ \frac{1}{|R|} \sum_{p \in R} \|\mathbf{v}_{pred}(p) - \bar{\mathbf{v}}_R\|_1 ]$, where $R$ is a superpixel region and $\bar{\mathbf{v}}_R$ is its mean feature.

\subsection{Rasterization-Native Semantic Extraction}
\label{sec:extraction}
To robustly extract semantics and boundaries, we process features directly in the 2D rendered space rather than via 3D primitives. First, the rasterizer projects 3D semantic features into a 2D feature map $\mathcal{F}_{sem}$, alongside the RGB image $\mathbf{I}_{rgb}$. To resolve visual-language ambiguity \cite{ref31} in dense point-wise matching, we introduce \textbf{Region-aware Similarity Aggregation}. 

An unsupervised segmentation algorithm decomposes the rendered image into superpixel regions $\Omega = \{R_1, \dots, R_K\}$. We aggregate the pixel-wise cosine similarity between $\mathcal{F}_{sem}$ and a target text embedding $\mathbf{v}_{text}$ via mean pooling within each $R_k$:
\begin{equation}
    S(R_k, \mathcal{T}) = \frac{1}{|R_k|} \sum_{p \in R_k} \left\langle \frac{\mathcal{F}_{sem}(p)}{\|\mathcal{F}_{sem}(p)\|_2}, \frac{\mathbf{v}_{text}}{\|\mathbf{v}_{text}\|_2} \right\rangle
\end{equation}
This suppresses spurious background activations and resolves local ambiguities. Assigning this aggregated score to all pixels in $R_k$ yields a coarse probability map $\mathbf{M}_{coarse}$. Finally, an edge-preserving Guided Filter $G$ uses $\mathbf{I}_{rgb}$ as structural guidance to recover precise pixel-level boundaries: $\mathbf{M}_{final} = G(\mathbf{M}_{coarse}, \mathbf{I}_{rgb})$.

\section{Experiments}

\subsection{Datasets and Evaluation Metrics}
To evaluate DeGenseGS in complex surgical environments featuring severe non-rigid deformations and topology-altering events such as electrocautery, we conduct experiments on five sequences from the CholecSeg8k \cite{ref29} benchmark and two sequences from the EndoVis18 \cite{ref30} benchmark. The mean Intersection over Union metric is adopted as the primary evaluation criterion for 3D semantic segmentation accuracy.

\subsection{Implementation Details}
We conduct all experiments on a single NVIDIA RTX 3090 GPU. The model is optimized using the Adam optimizer with an initial learning rate of $1.6 \times 10^{-3}$. We employ a coarse-to-fine schedule, freezing the deformation parameters for the first 3,000 iterations. Standard Gaussian densification and pruning are performed every 100 iterations. In terms of efficiency, DeGenseGS renders at $\sim$70 FPS, and a full text-prompted query with VLM matching and RNSE takes $\sim$0.56s per frame, on par with SurgTPGS \cite{ref15} ($\sim$67 FPS and $\sim$0.6s per frame).

\subsection{Quantitative and Qualitative Evaluation}
We compare DeGenseGS against state-of-the-art GS frameworks and vision-language integrated methods, including LangSplat \cite{ref13}, OpenGaussian \cite{ref24}, DGD \cite{ref25}, FE-4DGS \cite{ref26}, and SurgTPGS \cite{ref15}. To ensure a comprehensive assessment, variants equipped with different vision-language models such as SurgVLP \cite{ref27} and CAT-Seg \cite{ref28} are also evaluated.

\begin{table}[t]
\centering
\caption{Quantitative results evaluated by mean Intersection over Union on the CholecSeg8k benchmark. \textbf{Bold}: best in column; \underline{underline}: second best. Abbreviations include Abd. Wall for Abdominal Wall, Gallbl. for Gallbladder, L-hook for L-hook Electrocautery, Avg. for Average score, LS for LangSplat, and OG for OpenGaussian.}
\label{tab:cholecseg8k}
\resizebox{\textwidth}{!}{
\begin{tabular}{lcccccccccccccccc}
\toprule
\multirow{2}{*}{\textbf{Methods}} & 01\_00080 & \multicolumn{2}{c}{01\_00240} & \multicolumn{3}{c}{01\_15019} & \multicolumn{5}{c}{12\_15750} & \multicolumn{4}{c}{17\_01803} & \multirow{2}{*}{\textbf{Avg.}} \\
\cmidrule(lr){2-2} \cmidrule(lr){3-4} \cmidrule(lr){5-7} \cmidrule(lr){8-12} \cmidrule(lr){13-16}
 & Liver & Liver & Grasper & Abd. Wall & Grasper & Liver & Fat & Gallbl. & Grasper & L-hook & Liver & Abd. Wall & Fat & Grasper & Liver &  \\
\midrule
LangSplat & 64.87 & 57.91 & 4.50 & 23.84 & 4.13 & 24.94 & 25.63 & 3.42 & 0 & 5.97 & 25.03 & 43.38 & 12.31 & 4.13 & 10.43 & 20.35 \\
LS-SurgVLP & 65.03 & 57.14 & 4.31 & 32.79 & 5.61 & 27.45 & 4.10 & 3.34 & 0 & 5.97 & 22.59 & 42.97 & 10.30 & 4.53 & 10.34 & 19.50 \\
LS-CAT-Seg & 77.51 & 73.69 & 4.96 & 91.05 & 58.29 & \underline{39.87} & \textbf{75.83} & 5.93 & 0 & 23.81 & 30.60 & 72.93 & 24.00 & 14.46 & 12.84 & 39.37 \\
OpenGaussian & 2.64 & 0.08 & 0 & 0 & 0 & 0.42 & 0 & 0.68 & 0 & 8.05 & 0.21 & 0 & 1.63 & 0 & 0.19 & 0.93 \\
OG-SurgVLP & 1.76 & 0 & 0 & 0 & 0 & 0.57 & 0.19 & 0.60 & 0 & 0 & 0.15 & 13.02 & 1.42 & 2.24 & 0.70 & 1.38 \\
OG-CAT-Seg & 2.21 & 1.90 & 12.79 & 14.89 & 12.82 & 1.55 & 1.26 & 1.02 & 0 & 7.49 & 1.28 & 0 & 1.20 & 0 & 1.23 & 3.80 \\
DGD & 13.98 & 42.43 & 3.83 & 9.68 & 2.90 & 10.97 & 14.79 & 2.25 & 0.73 & 0 & 8.95 & 4.09 & 8.16 & 2.67 & 3.81 & 8.80 \\
FE-4DGS & 0.08 & 47.90 & 0.93 & 4.94 & 0.03 & 0.25 & 0.29 & 0.21 & 0 & 0 & 12.40 & 2.05 & 0.41 & 0.15 & 4.65 & 4.95 \\
SurgTPGS & \underline{88.70} & \underline{81.27} & \textbf{67.61} & \underline{97.18} & \textbf{68.57} & 37.42 & \underline{73.98} & \underline{7.63} & \underline{16.34} & \underline{54.04} & \underline{34.12} & \underline{89.26} & \underline{28.24} & \underline{58.74} & \underline{15.11} & \underline{53.46} \\
\midrule
\textbf{Ours} & \textbf{89.03} & \textbf{86.00} & \underline{61.05} & \textbf{97.66} & \underline{62.59} & \textbf{92.35} & 65.91 & \textbf{52.14} & \textbf{27.68} & \textbf{65.06} & \textbf{77.53} & \textbf{96.11} & \textbf{46.31} & \textbf{73.96} & \textbf{47.02} & \textbf{68.20} \\
\bottomrule
\end{tabular}
}
\bigskip
\caption{Quantitative results evaluated by mean Intersection over Union on the EndoVis18 benchmark. \textbf{Bold}: best in column; \underline{underline}: second best. Abbreviations include Avg. for Average score, LS for LangSplat, and OG for OpenGaussian.}
\label{tab:endovis18}
\resizebox{\textwidth}{!}{
\begin{tabular}{lcccccc}
\toprule
\multirow{2}{*}{\textbf{Methods}} & \multicolumn{2}{c}{Seq\_5} & \multicolumn{3}{c}{Seq\_9} & \multirow{2}{*}{\textbf{Avg.}} \\
\cmidrule(lr){2-3} \cmidrule(lr){4-6}
 & Inst-Wrist & Kidney-Parenchyma & Inst-Shaft & Inst-Wrist & Inst-Clasper &  \\
\midrule
LangSplat & 5.17 & 59.16 & 16.47 & 7.89 & 6.15 & 18.96 \\
LS-SurgVLP & 7.36 & 52.94 & 9.87 & 11.72 & 7.02 & 17.78 \\
LS-CAT-Seg & 17.16 & 54.42 & \textbf{38.49} & \underline{21.74} & 0.11 & 26.38 \\
OpenGaussian & 0 & 1.62 & 0 & 0 & 0 & 0.32 \\
OG-SurgVLP & 1.13 & 0.54 & 0 & 0 & 0 & 0.33 \\
OG-CAT-Seg & 0 & 0.06 & \underline{29.38} & 12.82 & 14.89 & 11.43 \\
DGD & 1.14 & 21.68 & 8.22 & 5.04 & 7.37 & 8.69 \\
FE-4DGS & 1.46 & 52.14 & 6.18 & 4.30 & 6.08 & 14.03 \\
SurgTPGS & \underline{43.29} & \underline{71.98} & 22.64 & 17.07 & \underline{40.48} & \underline{39.09} \\
\midrule
\textbf{Ours} & \textbf{52.96} & \textbf{82.22} & 20.26 & \textbf{22.24} & \textbf{45.46} & \textbf{44.63} \\
\bottomrule
\end{tabular}
}
\end{table}

As shown in Table \ref{tab:cholecseg8k} and Table \ref{tab:endovis18}, our method consistently outperforms all competitors. On CholecSeg8k, DeGenseGS establishes a new state-of-the-art score of 68.20\%, achieving a substantial 14.74\% absolute improvement over the best-performing baseline SurgTPGS at 53.46\%. Similarly, on EndoVis18, we achieve 44.63\%, significantly surpassing SurgTPGS at 39.09\%. In addition, averaged over all evaluated scenes from both datasets, DeGenseGS also improves the underlying RGB reconstruction quality over SurgTPGS by $+1.06$ in PSNR, $+0.030$ in SSIM, and $-0.0010$ in LPIPS; the segmentation gain ($+14.74\%$ mIoU on CholecSeg8k) is substantially larger than the reconstruction gain, indicating that the improvement primarily originates from the proposed geometry--semantic decoupling rather than from better RGB reconstruction alone.

\begin{figure}[t]
    \centering
    \includegraphics[width=\textwidth]{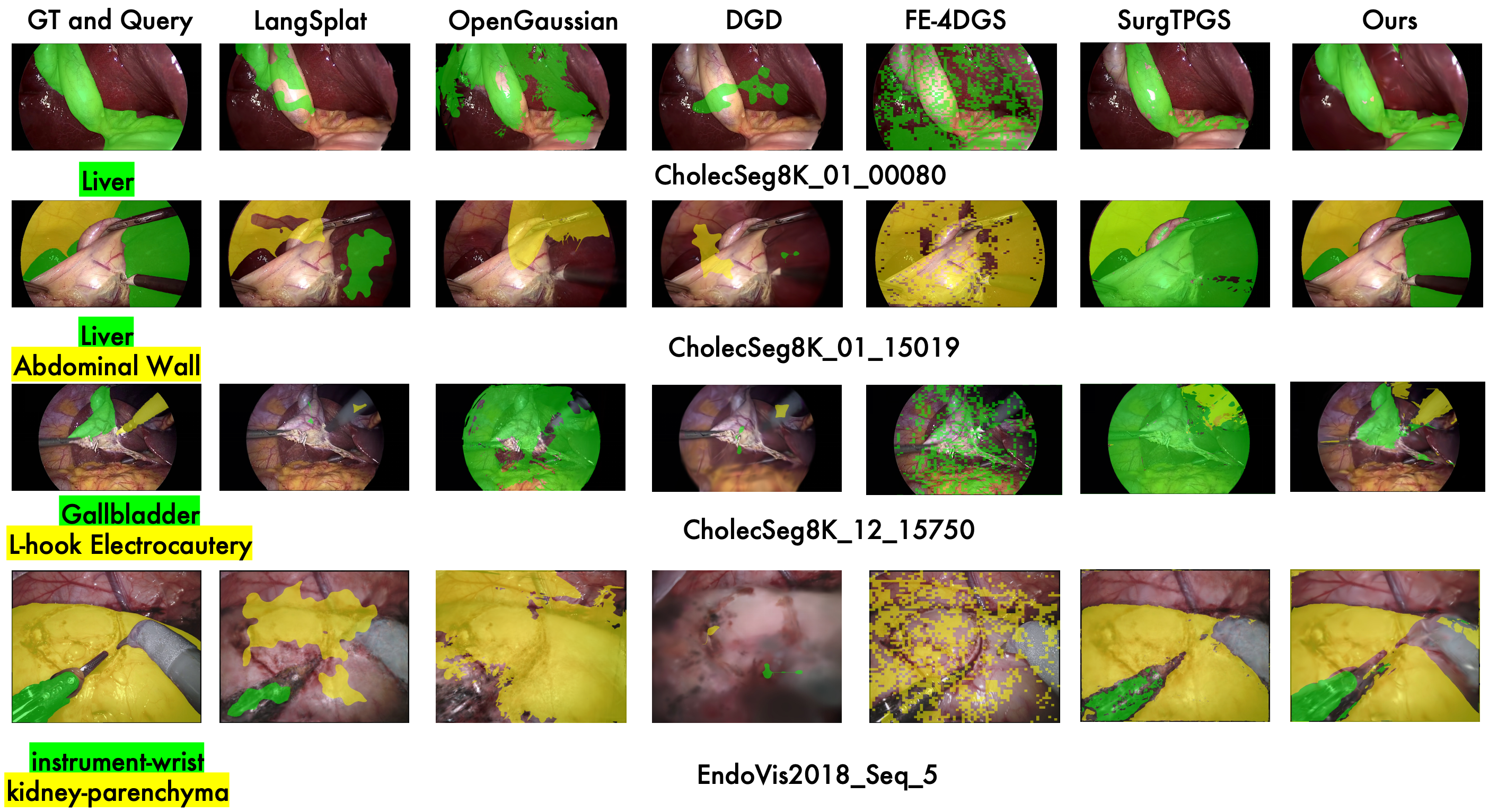}
    \caption{Qualitative result on CholecSeg8k and EndoVis18 datasets.}
    \label{fig:qualitative}
\end{figure}

This performance leap is particularly prominent in structures subject to severe topological and appearance alterations. For instance, gallbladder segmentation accuracy in sequence 12\_15750 surges from 7.63\% to 52.14\%, and instrument-wrist in Seq\_5 improves from 43.29\% to 52.96\%. These gains empirically validate that disentangling semantic evolution from geometric warping effectively prevents feature collapse, preserving high spatial coherence amidst surgical dynamics. Qualitatively, as illustrated in Fig. \ref{fig:qualitative}, DeGenseGS demonstrates significantly enhanced geometric connectivity and precise semantic boundaries.

\subsection{Ablation Study}

\begin{table}[t]

\centering

\caption{Ablation study of the proposed components. We report the mean Intersection over Union (mIoU \%) across the overall CholecSeg8k and EndoVis18 datasets, along with a specific subset (Seq 12\_15750) for fine-grained comparison.}

\label{tab:ablation}

\resizebox{\textwidth}{!}{

\begin{tabular}{cc ccc}

\toprule

\multirow{2}{*}{\textbf{S-G Decoupling}} & \multirow{2}{*}{\textbf{AASD}} & \multicolumn{2}{c}{\textbf{CholecSeg8k}} & \textbf{EndoVis18} \\

\cmidrule(lr){3-4} \cmidrule(lr){5-5}

 & & Seq 12\_15750 &Overall Average mIoU & Overall Average mIoU\\

\midrule

 $\times$ & $\times$ & 32.65 & 53.46 & 39.09 \\

 $\times$ & \checkmark & 34.12 & 55.12 & 40.21 \\

\checkmark & $\times$ & 46.21 & 66.35 & 43.45 \\

\checkmark & \checkmark & \textbf{48.57} & \textbf{68.20} & \textbf{44.63} \\

\bottomrule

\end{tabular}

}

\end{table}

To validate our contributions, Table \ref{tab:ablation} presents an ablation study evaluating our core components. Notably, our \textbf{Semantic-Geometric (S-G) Decoupling} encompasses both the Kinematics-Conditioned Latent Disentanglement (Sec. 2.2) and the Rasterization-Native Semantic Extraction (Sec. 2.4), which jointly decouple semantics from spatial warping at both the 4D feature level and the 2D rasterization level. For the rigidly coupled baseline ($\times$ for S-G Decoupling), we adopt the SurgTPGS architecture \cite{ref15}. 

As shown, while the Angular-Aligned Semantic Distillation (AASD) objective alone provides a marginal performance improvement by regularizing the VLM latent space, its overall impact remains constrained by the baseline's erroneous geometric warpings. In contrast, the complete S-G Decoupling framework acts as the primary driver of the substantial performance leap (e.g., boosting the CholecSeg8k average mIoU from 53.46\% to 66.35\%). This proves its crucial role in mitigating feature collapse and gradient-induced geometric artifacts during complex topology-altering events.

\section{Conclusion}
We present DeGenseGS, the first text-promptable 4D Gaussian Splatting framework capable of separately characterizing spatial deformation and semantic evolution for fine-grained surgical interaction. By implementing a dual-branch architecture for structural decoupling and rasterization-native semantic extraction, our method addresses the fundamental challenge of gradient interference between geometry and semantics. This decoupled paradigm prevents spurious geometric warping while ensuring spatially coherent surgical understanding. The superior performance validates DeGenseGS as a robust foundation for reliable perception in autonomous robotic-assisted surgery.

\begin{credits}
\subsubsection{\ackname}
This work was supported by Ministry of Education Tier 2 grant, Singapore (T2EP20224-0028), and Ministry of Education Tier 1 grant, Singapore (23-0651-P0001). This work was also supported by the China Scholarship Council under Grant No.\ 202506090084. Yimo Wang conducted this research as a CSC visiting PhD student at the National University of Singapore. 

\subsubsection{\discintname}
The authors have no competing interests to declare that are relevant to the content of this article.
\end{credits}

\bibliographystyle{splncs04}
\bibliography{Paper-1749}
\end{document}